\documentclass[11pt]{article}
\usepackage[preprint]{acl}
\usepackage{fontspec}
\usepackage{latexsym,amsmath,amssymb,booktabs,array,tabularx,graphicx,placeins,xcolor}
\newenvironment{promptblock}[1]{\par\medskip\noindent
\colorbox{black!7}{\parbox{\dimexpr\linewidth-2\fboxsep\relax}{\textbf{#1}}}
\par\nobreak\smallskip\begingroup\raggedright
\setlength{\parindent}{0pt}\setlength{\parskip}{3pt}
\setlength{\leftskip}{4pt}\setlength{\rightskip}{4pt plus 1fil}}
{\par\endgroup\nobreak\smallskip\hrule\medskip}
\hypersetup{pdftitle={From Belief Inference to Task Action: A Multi-Level Evaluation of Theory of Mind on Verifiable Tasks},pdfauthor={Zhenshuai Yin, Yongxin Guo, Mingda Zhang, Zhongze Luo, Haoyue Liu, Lianmin Zhang, Xiaoying Tang}}
\newcommand{\ind}{\mathbf{1}}
\newcommand{\iface}[1]{\textsf{#1}}
\newcolumntype{Y}{>{\raggedright\arraybackslash}X}
\title{From Belief Inference to Task Action:\\A Multi-Level Evaluation of Theory of Mind on Verifiable Tasks}
\author{
Zhenshuai Yin\textsuperscript{1}\quad
Yongxin Guo\textsuperscript{2}\quad
Mingda Zhang\textsuperscript{1}\quad
Zhongze Luo\textsuperscript{1}\\
\textbf{Haoyue Liu\textsuperscript{1}\quad
Lianmin Zhang\textsuperscript{3}\quad
Xiaoying Tang\textsuperscript{1}\thanks{Corresponding author.}}\\[3pt]
\textsuperscript{1}School of Science and Engineering,\\
The Chinese University of Hong Kong (Shenzhen),\\
Longgang, Shenzhen, Guangdong 518172, P.R. China\\
\textsuperscript{2}ByteDance Inc.\\
\textsuperscript{3}Shenzhen Research Institute of Big Data, Shenzhen, China
}

\date{}
\begin{document}
\maketitle
\begin{abstract}
Effective collaboration requires language models to understand what others know and believe, and to act accordingly. Models that assume others share their information may omit necessary communication and undermine coordination. Many theory-of-mind (ToM) benchmarks assess mental-state inference through story questions, but correct answers do not establish that models can use these judgments in subsequent tasks. We introduce a multi-level evaluation built on existing, verifiable tasks. We extend household collaboration and calendar scheduling with asymmetric information while retaining their original constraints and validators. Independent interfaces assess world understanding, information access, beliefs, decisions, and execution. Controls supply personal evidence or correct beliefs to distinguish state recovery from state use. Across six model configurations, these abilities diverge. On matched scheduling cases, Claude recovers all agent beliefs correctly but produces subjectively optimal plans in only 67.9\% of cases. In household collaboration, supplying beliefs raises Qwen-27B success from 55.0\% to 75.0\%, yet Qwen-4B answers every fixed supplied-belief source question correctly while completing only 20.0\% of the corresponding continuation condition. These results show why ToM evaluation must connect perspective inference to independently verified decisions and task consequences.
\end{abstract}

\section{Introduction}
Language-model theory of mind (ToM) is often tested through differences in what characters know. One person leaves before another moves an object, and the model must infer where the absent person believes it is. A conversation participant misses a message, and the model must distinguish their knowledge from that of the reader \citep{le2019,kim2023}. These designs test whether models preserve personal perspectives rather than simply repeat the latest fact.

\begin{figure}[t]
\centering
\includegraphics[width=\linewidth]{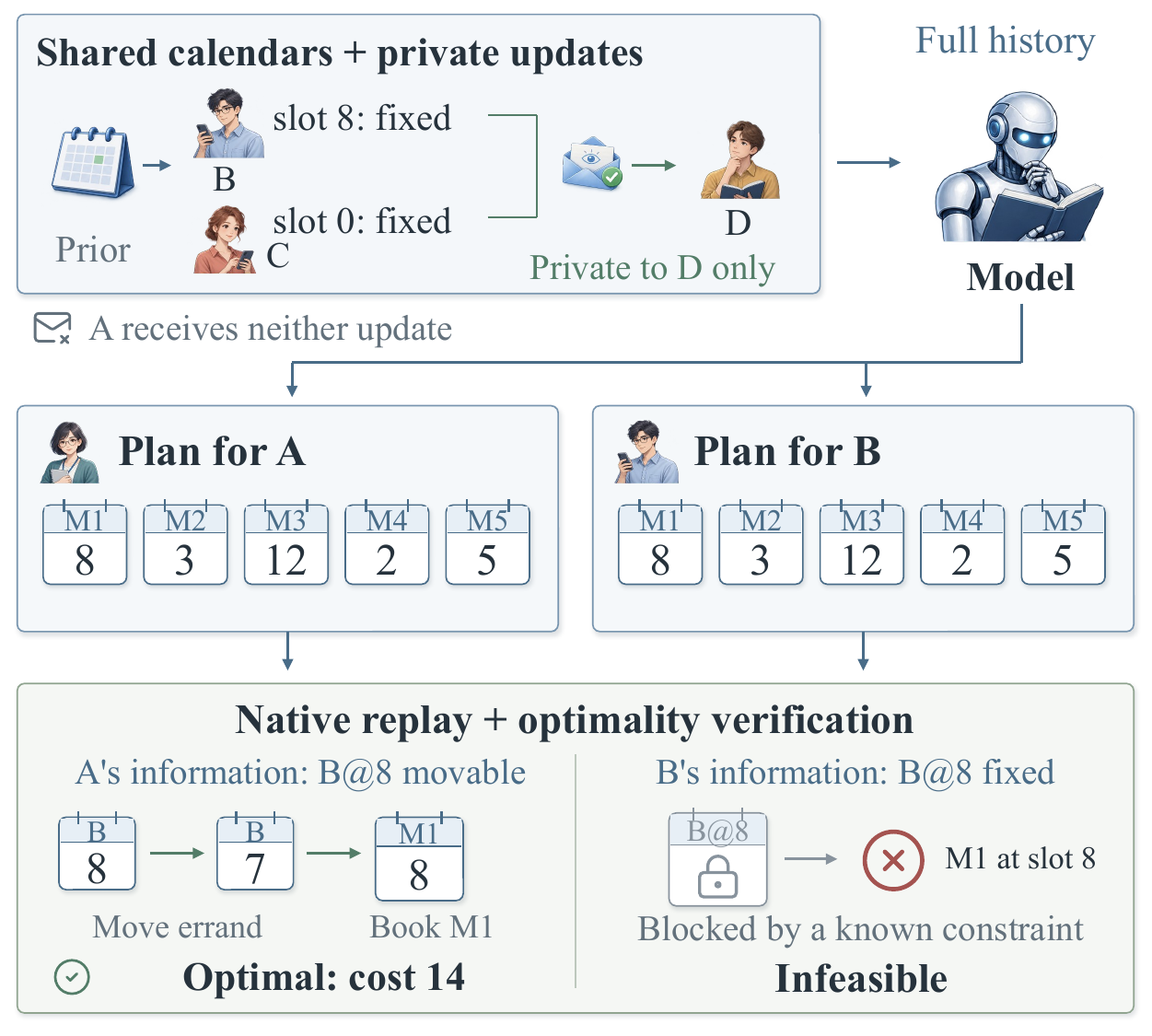}
\caption{An actual CalBench case connects private information to complete plans. The same five-meeting proposal is optimal for uninformed A but infeasible for B, who knows slot 8 is blocked. Both proposals fail in the actual world. Native replay and subjective solvers distinguish perspective-specific correctness from world success. Appendix~\ref{app:case} gives all five proposals.}
\label{fig:perspective}
\end{figure}

\begin{figure*}[t]
\centering
\includegraphics[width=\textwidth]{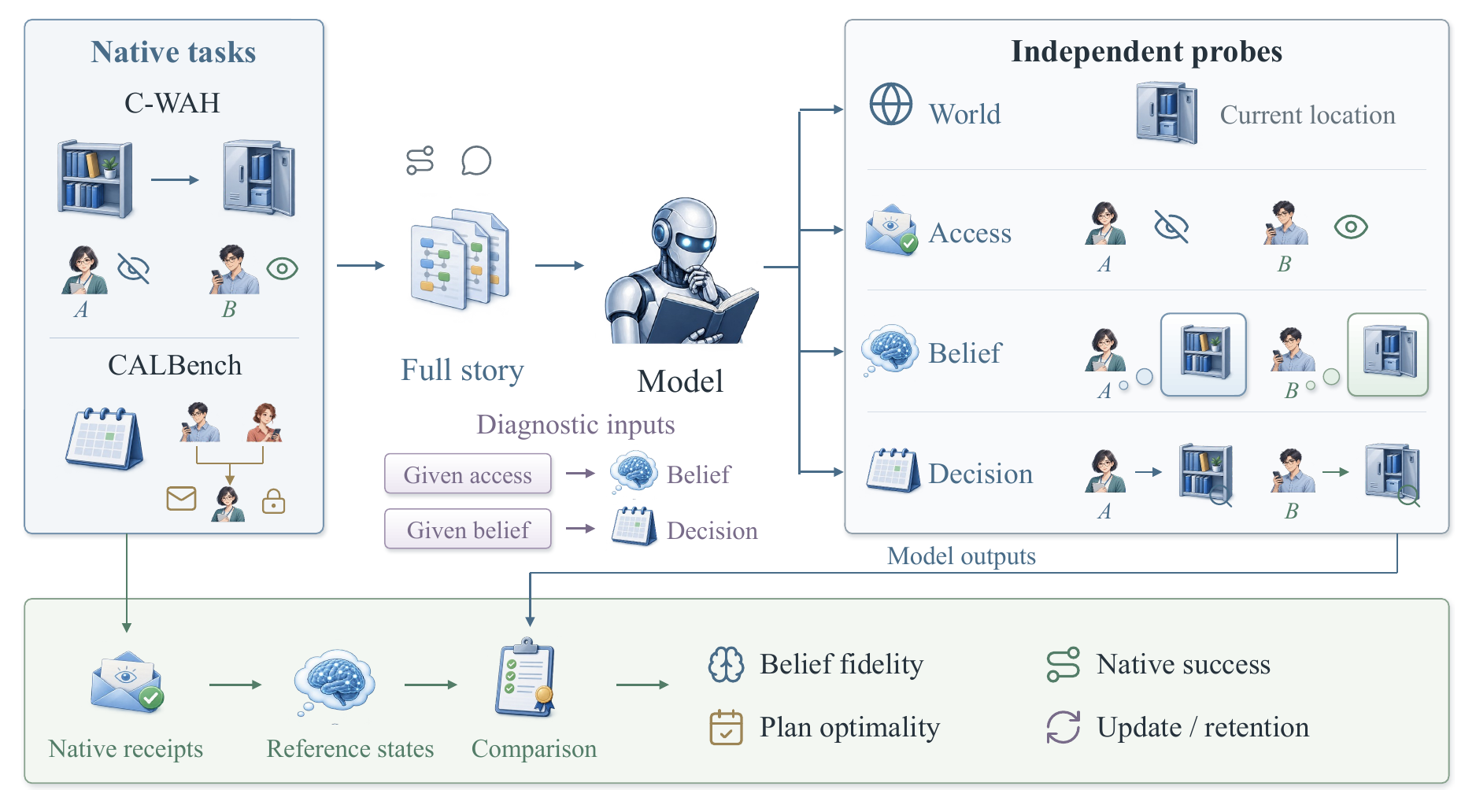}
\caption{Evaluation from personal evidence to task action. Independent interfaces probe world, access, beliefs, and decisions; native execution supplies action consequences and subsequent observations. Supplied-access and supplied-belief controls diagnose state recovery and use. The right-hand examples illustrate reference states, not measured model answers.}
\label{fig:architecture}
\end{figure*}

Belief-based action prediction extends this question to what a character will do \citep{gandhi2023}. Producing an entire constraint-satisfying plan, however, places a different demand on a model than predicting an action in a story. Functional ToM already studies adaptation to others and actual interaction \citep{riemer2025,juneja2026}. The remaining evaluation problem is to connect personal-state recovery, its use in decisions, and independently verifiable task outcomes without treating any one as a substitute for the others.

We ask whether models can recover what different agents know and use it within existing executable tasks. We extend C-WAH household collaboration \citep{zhang2024} and CalBench scheduling \citep{zou2026}, preserving their goals, constraints, and validators while constructing personal observation and communication histories. Figure~\ref{fig:perspective} illustrates the distinction: an uninformed person's plan can be correct under their evidence yet fail in the world. Conversely, a successful plan can happen to satisfy several perspectives without recovering all of them.

A single outcome cannot explain these cases. Failure may arise from misunderstanding events, misattributing evidence, losing an earlier belief, or planning incorrectly after the state is known. We connect five evaluation targets---world, access, belief, decision, and execution---and supply personal evidence or reference beliefs in diagnostic controls. Requests are independent, preventing errors in one answer from being mechanically propagated to the next.

Our contributions are a task-grounded ToM construction with native reference evidence and verifiable plans, a cross-domain protocol separating state recovery from state use, and a six-configuration analysis of their behavioral differences. The results expose both strong belief inference without optimal planning and near-perfect fixed decisions without successful long-horizon coordination.

\section{Related Work}
\paragraph{Mental-state inference.}
Cognitive AI connects beliefs and intentions to action understanding \citep{zhu2020}. ToMi controls story regularities in belief questions \citep{le2019}. FANToM tests beliefs, information accessibility, and answerability in multi-party conversations \citep{kim2023}. BigToM links perceptions, beliefs, desires, and actions through causal templates \citep{gandhi2023}. Hi-ToM examines higher-order reasoning \citep{wu2023}, while ToMBench and OpenToM broaden mental states and contextual diversity \citep{chen2024,xu2024}. ExploreToM searches programmatically tracked stories \citep{sclar2025}, and ToMATO constructs questions from role-playing conversations \citep{shinoda2025}. Percept-ToMi and Percept-FANToM isolate perception attribution \citep{jung2024}. We additionally evaluate complete plans under personal constraints and their execution.

\paragraph{Functional and interactive ToM.}
\citet{riemer2025} distinguish predicting opponents from using predictions in games. Collaborative bomb disposal combines beliefs with team outcomes \citep{li2023}. LLM-Coordination diagnoses environment understanding, ToM, and planning \citep{agashe2025}; instruction inference links intentions to plan feasibility and optimality \citep{saad2026}. SOTOPIA evaluates interactions with private social goals \citep{zhou2024}. MuMA-ToM studies observers' social reasoning about embodied agents \citep{shi2025}. Watch-And-Help and CoELA investigate household collaboration \citep{puig2021,zhang2024}, and EnactToM measures partner beliefs and task completion under controlled information \citep{juneja2026}. We combine dynamic interaction with solver-verified scheduling, comparing independent interfaces and supplied-state conditions on frozen tasks. Subjective correctness remains separate from actual success.

\paragraph{Verifiable task evaluation.}
PlanBench checks action sequences and state changes \citep{valmeekam2023}; ALFWorld connects textual decisions to embodied execution \citep{shridhar2021}; AgentBench spans interactive tasks \citep{liu2024}. AgentAD uses LLM agents to generate satellite-scheduling algorithms \citep{chen2026agentad}. CalBench adds distributed scheduling, private calendars, and displacement-cost optimization \citep{zou2026}. We retain native constraints and validators to assess perspective-dependent plans, distinguishing personal evidence from world knowledge.

\section{Problem Formulation}
We focus on first-order belief attribution and use under asymmetric information. Two agents can hold different beliefs about the same world because they obtain different evidence. A model reading their complete story must distinguish what it knows from what each agent knows, retain old states without new evidence, and use the resulting perspectives for decisions.

C-WAH and CalBench provide complementary tests. Household actions change the world and generate observations, allowing evaluation of belief maintenance during sequential coordination. Scheduling has explicit constraints and optimal costs, allowing complete plans to be checked under each person's information. Both provide task-based standards independent of the model's verbal explanation.

Our extensions retain the original environments, goals, and validators. C-WAH builds personal histories from actual routes, physical actions, and delivered messages. CalBench constructs initial calendars and subsequent public or private updates whose final calendars equal the frozen source task. Information conditions change which constraints each person knows, not the original task objective. The evaluation distinguishes recovering personal states, producing decisions correct under those states, and succeeding in the actual world.

This distinction separates predicting another person's informed choice from coordinating a successful joint action. A plan consistent with an outdated belief can reveal accurate perspective taking even when it fails physically. A coordinator can instead communicate missing information before assigning work. Silently replacing the person's belief with world truth tests neither operation. In our protocol, subjective planning checks whether the model respects the target perspective, while native collaboration tests whether it can use or change available information to advance a shared goal. Evaluating both prevents correct belief attribution from being confused with endorsement of an ineffective action.

\section{Evaluation Framework}
\subsection{Why Five Targets?}
The decomposition follows information dependencies. A model must understand what happened (\emph{world}), determine who obtained which evidence (\emph{access}), and integrate it over time (\emph{belief}). Evidence and belief differ because an older message can be superseded by observation, while lack of evidence requires retention. It must then use the resulting state to select an action or solve constraints (\emph{decision}). Actual execution can succeed, fail, or create new evidence (\emph{execution}). These targets locate behavioral errors, not five internal modules or a difficulty ordering.

\subsection{Evidence and Reference States}
Let $w_t$ be the world, $e_{1:t}$ the recorded events, and $R$ the observation and delivery rules. For agent $a$, references are
\begin{align}
H_{a,t}&=\operatorname{Access}(a,e_{1:t},R),\\
b_{a,t}&=\operatorname{Update}(b_{a,0},H_{a,t},R).
\end{align}
The model reads all events, whereas $H_{a,t}$ contains only evidence obtained by $a$. Reference states follow task rules rather than model-generated answers.

C-WAH records native physical graphs, personal observations, action feedback, and message delivery. Its symbolic room-level observations respect container and possession relations. Direct observation can override a conflicting message; a completed unsuccessful search can yield UNKNOWN. Without new evidence, a belief persists. Main histories contain physical events and public rules, not observer labels or reference beliefs. Rule replay verifies that the text supports the native references. Incidental encounters and observation-driven corrections remain valid events.

In CalBench, five agents broadcast their initial calendars. Two owners then learn that a movable commitment must remain in its slot. PUBLIC informs the group; PRIVATE informs a shared intermediary. Owners and actual recipients update, while uninformed agents retain previous calendars. Construction is deterministic from source tasks (Appendix~\ref{app:construction}).

Access and belief therefore have different reference semantics. In C-WAH, the access target retains positive location assertions a person has obtained, including assertions later superseded by observation. The belief target instead identifies the current state after those events. Remembering an old assertion is correct in the former and can be incorrect in the latter. Separate targets distinguish selecting evidence from resolving its temporal consequences.

\subsection{Independent Interfaces and Controls}
Seven shared interfaces request world/access, narrative beliefs, structured-history beliefs, direct decisions, decisions with textual beliefs, decisions with symbolic beliefs, and beliefs/decisions with supplied access. C-WAH adds spatial facts and public-map routes (Table~\ref{tab:interfaces}). Each request has an independent context and reference target.

Supplied access explicitly identifies personal evidence while retaining temporal integration. In C-WAH it provides deduplicated positive location assertions, not a complete ordered observation history; unsuccessful searches must still be interpreted from the narrative. In CalBench it provides ordered event identifiers. Supplied beliefs remove state-recovery demands but retain perspective-specific decisions. Narrative/structured histories and textual/symbolic beliefs test representation changes. Comparisons on the same stories distinguish attribution, maintenance, and constraint use without feeding predicted answers between interfaces.

These controls answer progressively narrower questions. If the relevant evidence is explicitly identified, can the model recover the target belief? If that belief is supplied correctly, can it choose an action appropriate to it? If physical movement succeeds, can it recover the perspectives produced by that movement? Independent requests make disagreements observable without an earlier incorrect answer determining every later input. Native continuation complements these controlled snapshots by allowing decisions to change the subsequent history. It measures whether the model can sustain coordination when information acquisition and action selection become its own responsibility.

\subsection{Task Decisions and Execution}
C-WAH fixed decisions specify an actor, object, and goal and ask for the source in that actor's belief. Native continuation permits assignments, messages, and exploration within the original task budget. Route interventions compare model-selected and supplied public-graph paths from matched native checkpoints, followed by a belief query scored against each resulting trajectory.

CalBench requires each proposer to schedule five meetings among five agents over sixteen slots. Entries are free, movable at a cost, or immovable. Every pair of meetings shares a participant, requiring distinct slots. A subjective solver minimizes displacement cost under the proposer's calendars. Native replay validates relocations and meetings. Independent optimization and replay must agree on reference costs; tied optima are accepted. The same proposal is checked in the actual world. A subjectively correct plan need not be world-feasible.

\section{Experimental Setup}
\paragraph{Configurations.}
We evaluate Qwen3.5-4B, 9B, and 27B with thinking disabled, plus Claude Opus 5.5, Gemini 3.1 Pro Preview, and GPT-6.1 Sol. Frozen inputs, references, and semantic scoring are shared. Qwen uses temperature 0, top-$p$ 1, and limits of 4096 output tokens for C-WAH and 12288 for CalBench. Closed services request medium reasoning and 32768 tokens with unspecified sampling controls. Resources and parameters appear in Appendix~\ref{app:interfaces}.

\paragraph{Tasks and common support.}
C-WAH has 43 configurations across seven houses, fixed seeds, 218 event-selected nodes, and nine interfaces: 1962 fixed requests per model. Direct and supplied-belief continuations each schedule 43 episodes; two route arms each schedule 36 configurations. Native technical coverage is 158/158 except 9B's 154/158. Six-model paired comparisons use 40 common continuation configurations and 35 route configurations. Episodes have a 250-step budget including preparation and at most 20 continuation decisions. Three deterministic scripts provide 129 shared reference episodes.

CalBench has 45 uniform-cost and 45 tiered-cost tasks, each with PUBLIC/PRIVATE stories and seven interfaces: 1260 requests per model. Complete coverage is 1136, 1015, and 1251 for Qwen-4B/9B/27B, and 1260 for each closed configuration. The main comparison uses 916 common complete request keys. Cross-interface analyses further match stories; public/private comparisons require both variants. All-valid results and per-interface coverage are separately reported.

Matching follows the question being compared. Within a calendar interface, every model answers the same complete request keys. A belief--planning comparison also requires both interfaces for the same story, and a PUBLIC/PRIVATE comparison requires both variants of the same source task. Consequently, these analyses have different denominators even though they begin with the same 90 tasks. Native comparisons similarly require both conditions to be technically valid in every configuration. This prevents differences in observed task composition from being mistaken for differences between models or input conditions.

\paragraph{Scoring.}
C-WAH uses English histories and scores person--object entries and full decisions. CalBench uses Chinese histories, requires all 80 entries for a correct agent belief, and all five meetings for a feasible, minimum-cost proposal. Timeouts/truncations receive bounded retries, adopting the first complete output without selecting by correctness. JSON extraction accepts fences and harmless surrounding text without changing content. Missing targets and semantic errors in complete answers remain incorrect. Incomplete outputs affect coverage, not capability scores. Nodes cluster within 43 configurations and calendar agents/stories within 90 source tasks. We report paired counts and configuration means, not significance from repeated entries.

\begin{table*}[t]
\centering\footnotesize
\setlength{\tabcolsep}{4pt}
\begin{tabularx}{\textwidth}{@{}p{.25\textwidth}>{\centering\arraybackslash}X>{\centering\arraybackslash}X>{\centering\arraybackslash}X>{\centering\arraybackslash}X>{\centering\arraybackslash}X>{\centering\arraybackslash}X>{\centering\arraybackslash}X@{}}
\toprule
\multicolumn{8}{@{}l}{\textbf{C-WAH: fixed nodes}} \\
\midrule
Model & World & Access & \shortstack{Narrative\\belief} & \shortstack{Struct.\\belief} & \shortstack{Given\\access\\belief} & \shortstack{Direct\\decision} & \shortstack{Given\\belief\\decision} \\
\midrule
Qwen3.5-4B & 93.1 & 33.8 & 49.3 & 38.6 & 60.2 & 66.5 & \textbf{100.0} \\
Qwen3.5-9B & 88.6 & 24.2 & 60.3 & 49.2 & 69.0 & 72.5 & 99.1 \\
Qwen3.5-27B & 96.3 & 42.3 & 62.8 & 63.6 & 91.4 & 82.1 & \textbf{100.0} \\
Claude Opus 5.5 & 99.0 & 75.8 & 71.2 & 73.4 & 92.3 & \textbf{87.6} & \textbf{100.0} \\
Gemini 3.1 Pro Preview & \textbf{99.4} & \textbf{99.4} & \textbf{99.4} & \textbf{99.0} & \textbf{99.9} & 86.2 & \textbf{100.0} \\
GPT-6.1 Sol & 95.0 & 86.5 & 94.7 & 92.0 & 99.6 & 83.9 & \textbf{100.0} \\
\bottomrule
\end{tabularx}
\par\medskip
\begin{tabularx}{\textwidth}{@{}p{.25\textwidth}>{\centering\arraybackslash}X>{\centering\arraybackslash}X>{\centering\arraybackslash}X>{\centering\arraybackslash}X>{\centering\arraybackslash}X>{\centering\arraybackslash}X>{\centering\arraybackslash}X@{}}
\toprule
\multicolumn{8}{@{}l}{\textbf{CalBench: common complete requests across models}} \\
\midrule
Model & World & Access & \shortstack{Narrative\\belief} & \shortstack{Struct.\\belief} & \shortstack{Direct\\planning} & \shortstack{Text belief\\planning} & \shortstack{Symbolic\\planning} \\
\midrule
Qwen3.5-4B & 23.8 & 6.3 & 3.2 & 3.8 & 13.2 & 30.6 & 35.2 \\
Qwen3.5-9B & 76.7 & 20.7 & 10.7 & 18.9 & 52.0 & 53.6 & 53.5 \\
Qwen3.5-27B & 99.9 & 81.5 & 96.2 & 96.2 & 88.7 & 85.9 & 91.2 \\
Claude Opus 5.5 & \textbf{100.0} & 97.2 & \textbf{100.0} & 97.7 & 68.9 & 77.0 & 81.3 \\
Gemini 3.1 Pro Preview & 99.2 & \textbf{99.0} & 97.0 & \textbf{99.4} & \textbf{100.0} & \textbf{100.0} & \textbf{100.0} \\
GPT-6.1 Sol & 99.9 & 98.3 & 92.8 & 96.0 & 81.0 & 79.7 & 80.7 \\
\bottomrule
\end{tabularx}
\par\medskip
\begin{tabularx}{\textwidth}{@{}p{.25\textwidth}>{\centering\arraybackslash}X>{\centering\arraybackslash}X>{\centering\arraybackslash}X>{\centering\arraybackslash}X>{\centering\arraybackslash}X>{\centering\arraybackslash}X@{}}
\toprule
\multicolumn{7}{@{}l}{\textbf{C-WAH: native tasks and post-action perspectives}} \\
\midrule
Model & \shortstack{Direct\\task success} & \shortstack{Given belief\\task success} & \shortstack{Model route\\entries} & \shortstack{Given route\\entries} & \shortstack{Model route\\whole table} & \shortstack{Given route\\whole table} \\
\midrule
Qwen3.5-4B & 12.5 & 20.0 & 54.6 & 53.0 & 0.0 & 0.0 \\
Qwen3.5-9B & 5.0 & 2.5 & 61.2 & 61.6 & 0.0 & 0.0 \\
Qwen3.5-27B & 55.0 & 75.0 & 62.0 & 61.0 & 2.9 & 0.0 \\
Claude Opus 5.5 & \textbf{95.0} & 92.5 & 93.7 & 68.8 & 71.4 & 0.0 \\
Gemini 3.1 Pro Preview & 92.5 & 97.5 & \textbf{100.0} & \textbf{100.0} & \textbf{100.0} & \textbf{100.0} \\
GPT-6.1 Sol & \textbf{95.0} & \textbf{100.0} & 90.9 & 91.4 & 62.9 & 65.7 \\
\bottomrule
\end{tabularx}
\par\medskip
\caption{Results on common support across six configurations (\%); bold marks column maxima. C-WAH uses 218 fixed nodes (1,426 world entries; 2,852 access/belief entries). CalBench uses 916 common complete requests, with interface denominators in the appendix. Native comparisons pair 40 continuation and 35 route configurations. Belief entries are pooled; whole tables must be entirely correct. Coverage and all-valid results are separate. Columns measure different targets.}
\label{tab:main-results}
\end{table*}

\section{Results and Analysis}
\subsection{Recovering Personal Evidence}
World comprehension and evidence attribution differ (Table~\ref{tab:main-results}). Qwen-4B recovers 93.1\% of household world entries but 33.8\% of personal access entries. Qwen-9B is better at narrative beliefs yet worse at access. Joint evidence confirms that this is not merely a difference in scoring units: Qwen-27B has 192 nodes with fully correct spatial answers but incomplete access answers. Gemini has only one. Personal attribution is a distinct, model-dependent requirement beyond understanding the physical scene.

Supplied access raises household belief accuracy from 49.3\% to 60.2\% for 4B, 60.3\% to 69.0\% for 9B, and 62.8\% to 91.4\% for 27B. It removes only part of the smaller configurations' difficulty: identifying relevant evidence is insufficient when integrating it into a complete state remains unreliable. Claude improves from 71.2\% to 92.3\%, whereas Gemini is near ceiling in both conditions.

\begin{figure}[!htb]
\centering
\includegraphics[width=\linewidth]{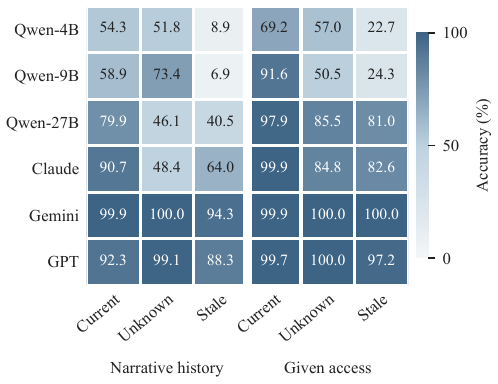}
\caption{Information boundaries on the same household references: 1446 current, 1159 unknown, and 247 stale entries. Narrative and supplied-access inputs test whether agents' states remain distinct from world truth. Nodes repeat within 43 configurations.}
\label{fig:cwah-states}
\end{figure}

Figure~\ref{fig:cwah-states} separates current facts from unknown and stale states. The latter require resisting the reader's knowledge. Explicit access thus concerns whose evidence controls a belief, not simply remembering more facts. Retention/update analyses further separate temporal demands (Appendix Figure~\ref{fig:cwah-dynamics}): Claude errs on 35.7\% of no-new-evidence entries but 7.6\% of updates, while GPT's rates are 4.4\% and 15.9\%. Their residual weaknesses differ despite both benefiting from supplied evidence.

\subsection{Using Beliefs in Decisions and Action}
\begin{figure}[!htb]
\centering
\includegraphics[width=\linewidth]{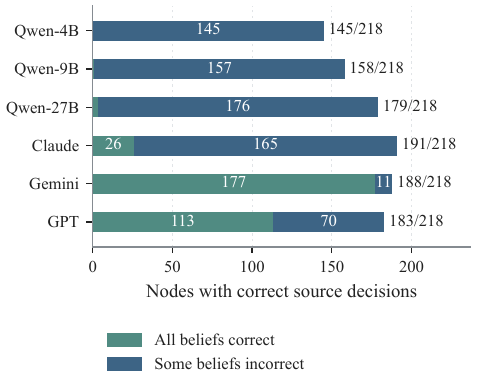}
\caption{Correct source choices partitioned by whether the independent narrative belief table is entirely correct, on the same 218 nodes. Correct choices coexist with partial state errors; a task-relevant answer does not certify the entire perspective.}
\label{fig:cwah-action}
\end{figure}

Fixed source selection is nearly solved with reference beliefs: five configurations achieve 218/218 and 9B achieves 216/218. Correct direct choices need not imply complete perspectives: all 145 correct 4B choices coexist with some incorrect belief entries, as do 157 of 158 for 9B (Figure~\ref{fig:cwah-action}). A decision may depend on only one correct entry. Conversely, Gemini has 26 nodes with a correct entire belief table but an incorrect independent source answer. State recovery and state use must both be tested.

The partition changes how success should be interpreted. For Gemini, 177 correct choices coincide with a fully correct belief table and only 11 with partial errors. For GPT, the corresponding counts are 113 and 70. The same successful decision can therefore accompany substantially different state recovery. These are paired observations at the same nodes, not an assumption that a correct choice requires every unrelated entry to be correct.

\begin{figure}[!htb]
\centering
\includegraphics[width=\linewidth]{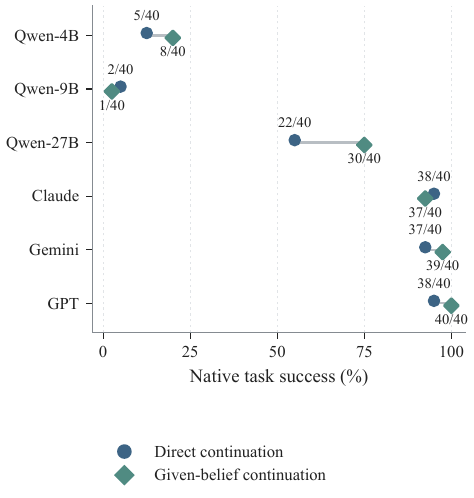}
\caption{Native task success on 40 common configurations. Circles denote direct continuation and diamonds supplied beliefs. Labels show successful episodes. Lines connect input conditions, not time.}
\label{fig:cwah-tasks}
\end{figure}

Native continuation makes the distinction consequential (Figure~\ref{fig:cwah-tasks}). Reference beliefs raise 27B success from 22/40 to 30/40 but 4B only from 5/40 to 8/40. For 9B, success remains 2/40 and 1/40 despite near-perfect fixed supplied-belief decisions. In each 9B continuation condition, 35/40 episodes issue no assignments. The model instead attempts extensive exploration (1658 and 1712 commands). Answering a specified source question does not establish the ability to select goals, turn information into assignments, or make progress within a decision budget. This is a policy-level observation, not evidence that larger models necessarily perform worse.

Closed configurations complete 37--40 paired episodes, but supplied beliefs do not uniformly help: Claude changes from 38 to 37. Shared scripts provide context on all 43 tasks: belief-first completes 42, world-first 13, and exploration-first 39. Both exploiting personal information and acquiring missing evidence can support collaboration; a true world location alone need not be a useful instruction for an uninformed executor.

\subsection{Tracking Perspectives After Action}
All supplied-route transports succeed on the 35 common configurations. Nevertheless, no 4B, 9B, or 27B post-action belief table is entirely correct in that condition. Their pooled-entry accuracies are 53.0\%, 61.6\%, and 61.0\%, respectively. Successful movement therefore does not demonstrate tracking of what each person learned.

Claude's configuration-mean belief accuracy is 93.9\% after model-selected routes but 69.0\% after supplied routes (Appendix Figure~\ref{fig:cwah-route}); Gemini is perfect in both. A route specifies movement, not automatically its informational consequences. Paths can induce different observations and histories. Native references expose this dependence while separating transport success from perspective recovery. The 4B model-selected arm also has three unsuccessful transports, which remain valid task failures.

\subsection{Planning Under Subjective Constraints}
\begin{figure}[!htb]
\centering
\includegraphics[width=\linewidth]{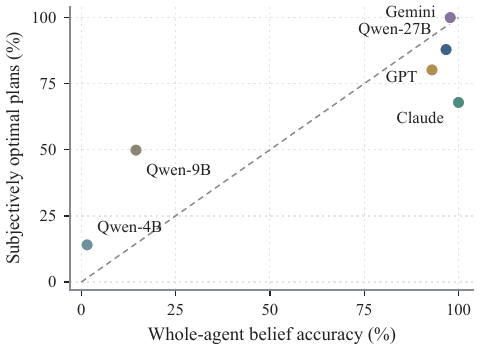}
\caption{Belief recovery versus planning on the same 91 stories (455 agent cases per axis). Beliefs require all 80 entries; plans require subjective feasibility and optimality. The dashed line marks equal percentages, not equivalent metrics.}
\label{fig:calbench-capability}
\end{figure}

On 91 stories complete for both narrative beliefs and direct planning across all models, Claude recovers 455/455 agent beliefs but produces only 309/455 optimal proposals (67.9\%). Qwen-27B recovers 440 beliefs and produces 400 optimal proposals. Qwen-9B produces 227 optimal proposals but only 66 complete beliefs. These differences do not yield one common ranking (Figure~\ref{fig:calbench-capability}). A plan can require fewer distinctions than an entire calendar state, while recovering that state does not ensure solving its constraints.

Supplied-state controls use 67 stories complete in all three planning interfaces and all six configurations. For 4B, correct plans rise from 44/335 directly to 107/335 with textual beliefs and 141/335 with symbolic beliefs. State support helps, but downstream difficulty remains. Claude rises from 232 to 259 and 283, whereas 9B scores 182, 177, and 179. Correct input states do not universally remove the limiting demand (Appendix Figure~\ref{fig:calbench-controls}). Gemini attains 335/335 in every condition, yet fails three direct household episodes: static optimization and dynamic coordination test different abilities.

Feasibility and optimality reveal a further difference hidden by this last comparison. On these same 335 proposals, 9B produces 192 feasible plans directly and 237 with textual beliefs. Yet its optimal count falls from 182 to 177: feasible but suboptimal plans increase from 10 to 60. Thus, a nearly unchanged optimality score conceals better constraint satisfaction alongside more failures to minimize cost. Claude's feasible counts rise from 270 to 302 under textual beliefs, but 43 of the latter remain suboptimal (Appendix Figure~\ref{fig:calbench-errors}). Supplied states can support compliance with known calendars without solving the remaining optimization problem. This distinction matters for interpreting ToM: violating a person's known constraints, finding an unnecessarily costly feasible plan, and using information they never received are different behavioral failures. A task validator makes the first two measurable rather than reducing every unsuccessful proposal to a wrong belief answer.

Public/private comparisons use 29 complete source-task pairs. Among them, 9B and GPT lose all five optimal proposals on seven and five tasks, whereas Gemini remains perfect (Appendix Figure~\ref{fig:calbench-communication}). The manipulation changes both who knows the constraints and how many subjective problems must be solved. Copying world calendars illustrates the need for targeted metrics: on common PRIVATE belief cases it scores 98.5\% of entries but only 20.0\% of complete agent states and zero stale entries. High ordinary entry accuracy can conceal failure to preserve another's perspective.

Planning scores also require a perspective-sensitive interpretation. Reusing the solver's world-optimal plan for every agent is subjectively optimal in 466/485 common direct-planning cases (96.1\%). Only 19 cases from six source tasks have strictly lower subjective than world-optimal cost. On these cases, Qwen-4B/9B/27B, Claude, Gemini, and GPT score 0, 0, 13, 10, 19, and 10 out of 19, respectively (Appendix~\ref{app:planning-overlap}). This privileged solver reference measures overlap between scoring targets, not model behavior or an equal-compute baseline. Aggregate planning success alone therefore does not establish use of a personal perspective.

\section{Discussion}
Task-grounded evaluation reveals two gaps. Understanding events does not ensure attributing evidence and maintaining personal beliefs. A correct state or fixed decision does not ensure effective planning and coordination. The 4B/9B results make the latter visible: fixed supplied-belief decisions are nearly perfect, but actual assignments and task completion remain weak. Closed configurations also separate calendar-state recovery from optimization and physical success from post-action beliefs.

These patterns imply different support needs. Explicit personal evidence can assist attribution but cannot replace temporal integration. Reference beliefs remove state uncertainty, but models must still preserve subjective constraints, choose goals, and execute productive commands. A single ToM or success score cannot identify which support is needed. Independent state questions, perspective-sensitive plans, and native consequences provide complementary evidence.

The two domains also clarify when information gathering belongs to the task. In scheduling, each proposer is evaluated under a fixed personal calendar: adding an unreceived update changes the problem being solved. In household continuation, communication and exploration are available actions that can legitimately change what an executor knows. The coordinator must decide when current information supports an assignment and when obtaining or transmitting evidence is useful. The common requirement is to respect the information available at the time of the decision, not to keep an agent uninformed indefinitely. This explains why passive belief questions and active coordination should be complementary tests rather than competing definitions of ToM.

For evaluation design, the unit of success should match the claim. Entry accuracy describes local state recovery, whole-agent accuracy tests a complete perspective, subjective feasibility checks known constraints, and optimality checks efficient use of those constraints. Native outcomes then measure whether the resulting policy reaches the shared goal. Reporting these targets together exposes both overinterpretations: a successful action does not establish complete mental-state understanding, and a correct mental-state answer does not establish competent collaboration.

\section{Conclusion}
We extend verifiable household and scheduling tasks with asymmetric information, retaining native constraints and independent references. Across six configurations, belief inference, use in decisions, and task success diverge in both directions. Evaluating collaborative ToM requires checking not only what a model says others believe, but also the plans it constructs from those beliefs and the consequences of executing them. Our controls make the distinction testable on the same source tasks: supplying evidence probes state recovery, supplying beliefs probes state use, and native execution checks whether either translates into task progress.

\FloatBarrier
\section*{Limitations}
The evaluation covers six service configurations, 43 household configurations across seven houses, and 90 calendar tasks. Each uses one fixed seed without independent repeated runs. Reasoning budgets, sampling controls, and serving hardware differ. Results establish neither pure parameter-scaling effects nor statistical significance across independent runs.

\section*{Ethical Considerations}
We use simulated households and controlled calendars, without new human-subject experiments or personal calendar collection. Synthetic messages do not confer real privacy authorization. Task scores cannot establish reliability in real collaboration or replace consent and verification. Resource provenance and licensing are described in Appendix~\ref{app:resources}.

\appendix
\raggedbottom
\setcounter{dbltopnumber}{2}
\section{A Complete Perspective Example}
\label{app:example}
This example illustrates perspective constraints. It is neither a sampled CalBench task nor a model response. Formal experiments use five agents and total displacement cost across five meetings.

\paragraph{Complete prompt.}
``A, B, and C need to hold a one-hour meeting attended by all three. Candidate start times are 14:00, 15:00, and 16:00. Each agent must choose the earliest feasible slot according to the information they have obtained. All agents remember their information accurately and retain earlier judgments unless they receive new evidence. Each immediately knows changes to their own calendar. Private messages are known only to their sender and specified recipient. Other people do not automatically learn their contents.

Initially, A, B, and C confirm in a shared group conversation that everyone is free at 14:00, 15:00, and 16:00.

Next, B learns that they are unavailable at 14:00. B privately messages only A: `My calendar has changed. I cannot attend at 14:00.'

Then C learns that they are unavailable at 15:00. C privately messages only A: `My calendar has changed. I cannot attend at 15:00.'

There are no other messages, relays, or observations. You can read the complete story, but must answer separately using only the information A, B, and C have obtained. Return JSON with a \texttt{proposals} object whose keys are A, B, and C and whose values are their proposed earliest feasible start times. Return only this JSON.''

\paragraph{Reference output.}
\begin{quote}\small
\texttt{\{"proposals":\{}\\
\texttt{~~"A":"16:00",}\\
\texttt{~~"B":"15:00",}\\
\texttt{~~"C":"14:00"\}\}}
\end{quote}
A receives both updates, while B and C each know only their own change. Only 16:00 is feasible in the real world. Choosing 16:00 for everyone solves the global scheduling problem but does not select the subjectively earliest slot under B's and C's local knowledge.

\FloatBarrier
\section{Layer-wise Results and Belief Categories}
Tables~\ref{tab:cwah-counts} and~\ref{tab:calbench-counts} give the numerators and denominators behind the main table. Table~\ref{tab:beliefstates} further separates C-WAH current, unknown, and stale reference beliefs. Configurations use the same frozen references. An unknown state is a target to predict correctly, not a missing sample.
\begin{table*}[tp]
\centering\small
\begin{tabularx}{\textwidth}{@{}Yrrrrrr@{}}
\toprule
Metric & Qwen-4B & Qwen-9B & Qwen-27B & Claude & Gemini & GPT \\
\midrule
World & 1327/1426 & 1263/1426 & 1373/1426 & 1412/1426 & 1418/1426 & 1355/1426 \\
Access & 964/2852 & 690/2852 & 1207/2852 & 2163/2852 & 2836/2852 & 2466/2852 \\
Narrative beliefs & 1407/2852 & 1719/2852 & 1790/2852 & 2031/2852 & 2836/2852 & 2701/2852 \\
Structured beliefs & 1102/2852 & 1404/2852 & 1813/2852 & 2094/2852 & 2823/2852 & 2624/2852 \\
Given access: beliefs & 1717/2852 & 1969/2852 & 2606/2852 & 2631/2852 & 2850/2852 & 2840/2852 \\
Direct decisions & 145/218 & 158/218 & 179/218 & 191/218 & 188/218 & 183/218 \\
Given beliefs: decisions & 218/218 & 216/218 & 218/218 & 218/218 & 218/218 & 218/218 \\
Spatial facts & 4159/4431 & 4146/4431 & 4406/4431 & 4359/4431 & 4431/4431 & 4408/4431 \\
Public-map route & 186/218 & 187/218 & 218/218 & 218/218 & 218/218 & 218/218 \\
Given symbolic beliefs & 216/218 & 217/218 & 218/218 & 218/218 & 218/218 & 218/218 \\
\bottomrule
\end{tabularx}
\caption{Exact C-WAH counts on the same 218 nodes. Textual and symbolic supplied beliefs are scored separately.}
\label{tab:cwah-counts}
\end{table*}

\begin{table*}[tp]
\centering\small
\begin{tabularx}{\textwidth}{@{}Yrrrrrr@{}}
\toprule
Metric & Qwen-4B & Qwen-9B & Qwen-27B & Claude & Gemini & GPT \\
\midrule
World & 2281/9600 & 7365/9600 & 9592/9600 & 9598/9600 & 9520/9600 & 9592/9600 \\
Access & 38/600 & 124/600 & 489/600 & 583/600 & 594/600 & 590/600 \\
Narrative beliefs & 26/825 & 88/825 & 794/825 & 825/825 & 800/825 & 766/825 \\
Structured beliefs & 33/875 & 165/875 & 842/875 & 855/875 & 870/875 & 840/875 \\
Direct planning & 64/485 & 252/485 & 430/485 & 334/485 & 485/485 & 393/485 \\
Text beliefs: planning & 225/735 & 394/735 & 631/735 & 566/735 & 735/735 & 586/735 \\
Symbolic beliefs: planning & 243/690 & 369/690 & 629/690 & 561/690 & 690/690 & 557/690 \\
Given access: beliefs & 14/370 & 22/370 & 365/370 & 359/370 & 365/370 & 332/370 \\
Given access: planning & 10/370 & 118/370 & 338/370 & 290/370 & 365/370 & 240/370 \\
\bottomrule
\end{tabularx}
\caption{Exact CalBench counts on common complete support, corresponding to the main table. Supplied access is listed separately because it jointly requests beliefs and proposals, unlike the separate belief and planning interfaces.}
\label{tab:calbench-counts}
\end{table*}

\begin{table*}[tp]
\centering\small
\begin{tabularx}{\textwidth}{@{}Yrrrrrr@{}}
\toprule
Metric & Qwen-4B & Qwen-9B & Qwen-27B & Claude & Gemini & GPT \\
\midrule
Narrative: Current & 785/1446 & 851/1446 & 1156/1446 & 1312/1446 & 1444/1446 & 1335/1446 \\
Narrative: Unknown & 600/1159 & 851/1159 & 534/1159 & 561/1159 & 1159/1159 & 1148/1159 \\
Narrative: Stale & 22/247 & 17/247 & 100/247 & 158/247 & 233/247 & 218/247 \\
Structured: Current & 885/1446 & 937/1446 & 1240/1446 & 1222/1446 & 1426/1446 & 1317/1446 \\
Structured: Unknown & 200/1159 & 450/1159 & 466/1159 & 738/1159 & 1159/1159 & 1109/1159 \\
Structured: Stale & 17/247 & 17/247 & 107/247 & 134/247 & 238/247 & 198/247 \\
Given access: Current & 1000/1446 & 1324/1446 & 1415/1446 & 1444/1446 & 1444/1446 & 1441/1446 \\
Given access: Unknown & 661/1159 & 585/1159 & 991/1159 & 983/1159 & 1159/1159 & 1159/1159 \\
Given access: Stale & 56/247 & 60/247 & 200/247 & 204/247 & 247/247 & 240/247 \\
\bottomrule
\end{tabularx}
\caption{Exact C-WAH counts by reference belief category, using the same frozen entries across configurations.}
\label{tab:beliefstates}
\end{table*}

\begin{table*}[tp]
\centering\small
\begin{tabularx}{\textwidth}{@{}lYY@{}}
\toprule
Interface & Input in addition to task rules & Output target \\
\midrule
\iface{WORLD\_ACCESS} & Complete narrative & World state and personal evidence collections \\
\iface{BELIEF\_NL} & Complete narrative & Each agent's beliefs \\
\iface{BELIEF\_STRUCT} & Structured event history & Each agent's beliefs \\
\iface{GIVEN\_NL} & World state and textual reference beliefs & Decisions under personal information \\
\iface{GIVEN\_SYMBOL} & World state and symbolic reference beliefs & Decisions under personal information \\
\iface{DIRECT} & Complete narrative & Decisions under personal information \\
\iface{ACCESS\_GIVEN} & Complete narrative and reference access & Beliefs and decisions under personal information \\
\bottomrule
\end{tabularx}
\caption{Shared interfaces. C-WAH decisions specify the believed source for a fixed assignment. CalBench decisions give a complete five-meeting proposal for each of five agents. Supplied states come from the evaluator, not model answers to other interfaces.}
\label{tab:interfaces}
\end{table*}

\begin{figure}[!htb]
\centering
\includegraphics[width=\linewidth]{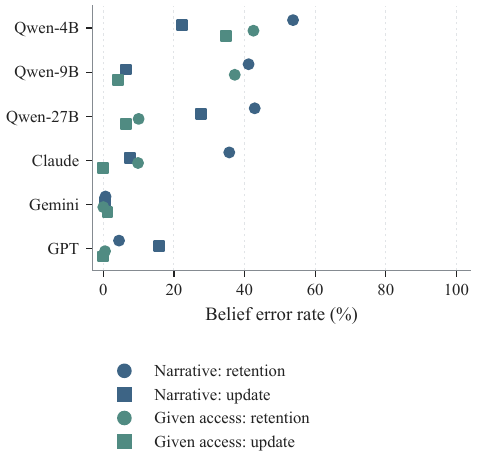}
\caption{Belief retention and updating across C-WAH models. Circles represent 2248 entries that should be retained without new evidence, and squares 170 actual updates. Blue denotes narrative input and green supplied access. The remaining 434 confirmation entries are excluded.}
\label{fig:cwah-dynamics}
\end{figure}

\section{Model Settings and Output Targets}
\label{app:interfaces}
Table~\ref{tab:parameters} records the actual request settings. Open and closed configurations share tasks and scoring but differ in reasoning controls. Unspecified sampling parameters are handled by each service and are not equivalent to explicit deterministic decoding.

\begin{table*}[tp]
\centering\small
\begin{tabularx}{\textwidth}{@{}Yllll@{}}
\toprule
Configuration & Reasoning control & C-WAH limit & CalBench limit & Temperature / top-$p$ \\
\midrule
Qwen3.5-4B, 9B & Thinking off & 4096 & 12288 & 0 / 1 \\
Qwen3.5-27B & Thinking off & 4096 & 12288 & 0 / 1 \\
Claude Opus 5.5 & Medium & 32768 & 32768 & Unspecified \\
Gemini 3.1 Pro Preview & Medium & 32768 & 32768 & Unspecified \\
GPT-6.1 Sol & Medium & 32768 & 32768 & Unspecified \\
\bottomrule
\end{tabularx}
\caption{Evaluated configurations and output limits in tokens. Closed configurations request the recorded model names and medium reasoning through a compatible API. Equal reasoning computation across services is not assumed.}
\label{tab:parameters}
\end{table*}

\paragraph{C-WAH.}
The opening specifies house connectivity, relevant objects and containers, goal counts, initial state, observation rules, and an independent teaching example. Subsequent history contains actual actions, routes, and communication. Source questions retain the assigned executor, object identifier, and goal, asking for the believed source. Main histories contain neither observer labels nor reference belief tables. Structured conditions encode physical events only. Every executed action, feedback item, and message enters subsequent continuation contexts.

\paragraph{CalBench.}
The opening defines calendar semantics, displacement costs, meeting participants, message delivery, and retention of earlier beliefs without updates. Beliefs are organized by holder, calendar owner, and slot. Proposals are organized by proposer and meeting identifier. Supplied-belief conditions show both world calendars and correct subjective calendars, explicitly stating that world truth does not automatically correct an agent's beliefs. Identical subjective calendars may be grouped, but outputs still list a complete plan for each of the five agents.

\paragraph{Answer extraction and semantic scoring.}
The common extractor retrieves existing JSON targets from complete responses, accepting fences, surrounding explanations, and separate thinking sections. For CalBench, defects in the work field or outer packaging do not incur an error when the complete answer can be extracted. Missing calendars, meetings, or states are not supplied. Missing target structures and wrong content are recorded separately. Agent belief correctness requires all 80 entries. Scheduling requires five valid meeting slots, native feasibility, and minimum subjective cost.

\section{Complete Coverage and Source-task Aggregation}
All six configurations complete 1962 fixed C-WAH requests. Native technical coverage is 158/158 for five configurations and 154/158 for Qwen-9B. Accepted jobs have sufficient inputs, and accepted route jobs match their handovers. Strict message/no-message pairs were not scheduled. Main native comparisons restrict to 40 configurations with both continuation conditions valid in all models and 35 with both route arms valid. Table~\ref{tab:native-all} reports all valid jobs, including configurations outside this intersection.

CalBench complete counts are 1136, 1015, and 1251 for Qwen-4B/9B/27B and 1260 for each closed configuration. The 916 common keys align models within each interface. Table~\ref{tab:full-coverage} separately reports all complete outputs, and Table~\ref{tab:interface-coverage} gives interface coverage. Table~\ref{tab:paired} requires all five agents to be correct in both stories of a source task. Ten agent instances do not replace one source task.

\begin{table*}[tp]
\centering\small
\begin{tabularx}{\textwidth}{@{}Yrrrrrr@{}}
\toprule
Metric & Qwen-4B & Qwen-9B & Qwen-27B & Claude & Gemini & GPT \\
\midrule
Complete-output coverage & 1136/1260 & 1015/1260 & 1251/1260 & 1260/1260 & 1260/1260 & 1260/1260 \\
Narrative: complete beliefs & 26/895 & 88/830 & 864/900 & 895/900 & 875/900 & 836/900 \\
Structured: complete beliefs & 33/900 & 165/875 & 867/900 & 880/900 & 895/900 & 865/900 \\
Direct planning & 107/815 & 281/545 & 768/895 & 582/900 & 900/900 & 669/900 \\
Text beliefs: planning & 255/855 & 424/785 & 759/885 & 688/900 & 900/900 & 712/900 \\
Symbolic beliefs: planning & 281/810 & 400/775 & 801/890 & 713/900 & 900/900 & 729/900 \\
Given access: planning & 18/765 & 133/415 & 733/885 & 654/900 & 890/900 & 564/900 \\
\bottomrule
\end{tabularx}
\caption{CalBench scores and coverage over all complete outputs. The main table compares the common support of all six configurations.}
\label{tab:full-coverage}
\end{table*}

\begin{table*}[tp]
\centering\small
\begin{tabularx}{\textwidth}{@{}Yrrrrrr@{}}
\toprule
Metric & Qwen-4B & Qwen-9B & Qwen-27B & Claude & Gemini & GPT \\
\midrule
Narrative beliefs & 0/76 & 0/76 & 70/76 & 76/76 & 73/76 & 60/76 \\
Structured beliefs & 0/85 & 2/85 & 79/85 & 81/85 & 84/85 & 72/85 \\
Direct planning & 2/29 & 10/29 & 23/29 & 9/29 & 29/29 & 17/29 \\
Text beliefs: planning & 6/63 & 18/63 & 44/63 & 33/63 & 63/63 & 41/63 \\
Symbolic beliefs: planning & 5/53 & 15/53 & 40/53 & 35/53 & 53/53 & 34/53 \\
\bottomrule
\end{tabularx}
\caption{Strict CalBench source-task accuracy. All five agents must be correct in both PUBLIC and PRIVATE. Denominators count common source tasks with complete outputs in both conditions.}
\label{tab:paired}
\end{table*}

\begin{table*}[tp]
\centering\small
\begin{tabularx}{\textwidth}{@{}Yrrrrrr@{}}
\toprule
Metric & Qwen-4B & Qwen-9B & Qwen-27B & Claude & Gemini & GPT \\
\midrule
WORLD\_ACCESS & 128/180 & 170/180 & 180/180 & 180/180 & 180/180 & 180/180 \\
BELIEF\_NL & 179/180 & 166/180 & 180/180 & 180/180 & 180/180 & 180/180 \\
BELIEF\_STRUCT & 180/180 & 175/180 & 180/180 & 180/180 & 180/180 & 180/180 \\
GIVEN\_NL & 171/180 & 157/180 & 177/180 & 180/180 & 180/180 & 180/180 \\
GIVEN\_SYMBOL & 162/180 & 155/180 & 178/180 & 180/180 & 180/180 & 180/180 \\
DIRECT & 163/180 & 109/180 & 179/180 & 180/180 & 180/180 & 180/180 \\
ACCESS\_GIVEN & 153/180 & 83/180 & 177/180 & 180/180 & 180/180 & 180/180 \\
\bottomrule
\end{tabularx}
\caption{Complete-output coverage by CalBench interface (180 scheduled per interface). Incomplete outputs are excluded from capability scores.}
\label{tab:interface-coverage}
\end{table*}

\begin{table*}[tp]
\centering\small
\begin{tabularx}{\textwidth}{@{}Yrrrrrr@{}}
\toprule
Metric & Qwen-4B & Qwen-9B & Qwen-27B & Claude & Gemini & GPT \\
\midrule
Native valid coverage & 158/158 & 154/158 & 158/158 & 158/158 & 158/158 & 158/158 \\
\shortstack{Direct\\task success} & 5/43 & 2/42 & 23/43 & 40/43 & 39/43 & 40/43 \\
\shortstack{Given belief\\task success} & 8/43 & 1/41 & 32/43 & 39/43 & 41/43 & 42/43 \\
\shortstack{Model route\\entries} & 264/484 & 295/484 & 299/484 & 453/484 & 484/484 & 440/484 \\
\shortstack{Given route\\entries} & 256/484 & 292/474 & 294/484 & 333/484 & 484/484 & 442/484 \\
\shortstack{Model route\\whole table} & 0/36 & 0/36 & 1/36 & 25/36 & 36/36 & 22/36 \\
\shortstack{Given route\\whole table} & 0/36 & 0/35 & 0/36 & 0/36 & 36/36 & 23/36 \\
Transport: model route & 33/36 & 36/36 & 36/36 & 36/36 & 36/36 & 36/36 \\
Transport: given route & 36/36 & 35/35 & 36/36 & 36/36 & 36/36 & 36/36 \\
\bottomrule
\end{tabularx}
\caption{Exact counts over all technically valid native jobs. Coverage differs by configuration; the main table uses paired common configurations.}
\label{tab:native-all}
\end{table*}

\section{Task Semantics of Planning Errors}
\label{app:errors}
\begin{figure*}[!t]
\centering
\includegraphics[width=.97\textwidth]{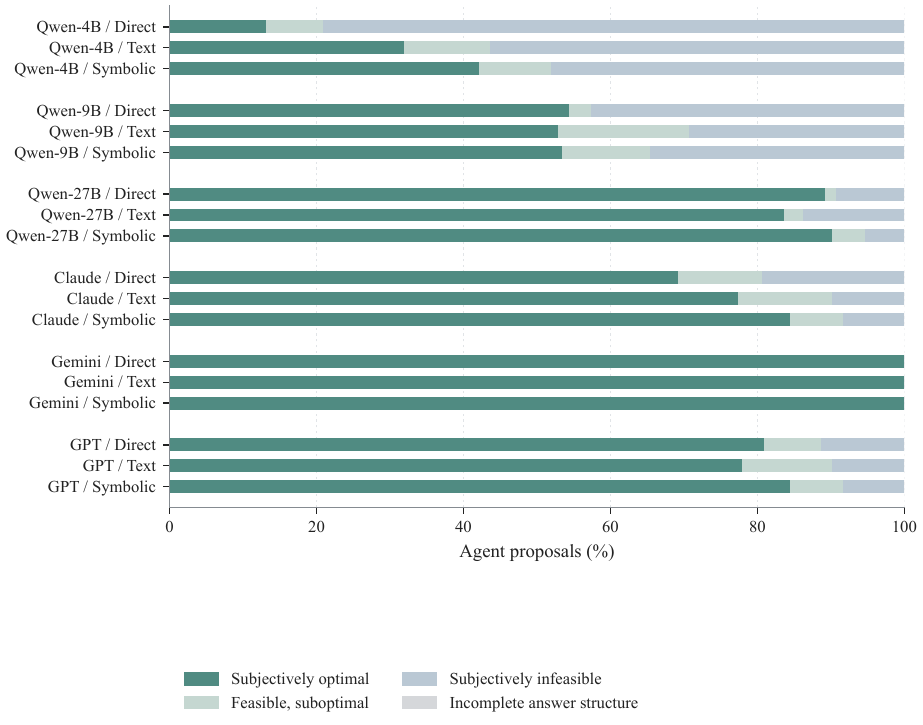}
\caption{Planning outcomes on matched stories. Direct, textual-belief, and symbolic-belief rows each contain 335 proposals. Incomplete structure means the original answer lacks a scorable plan, excluding recoverable fences and surrounding prose.}
\label{fig:calbench-errors}
\end{figure*}

Figure~\ref{fig:calbench-errors} decomposes outcomes for 67 stories complete across three planning interfaces and six configurations. Subjective optimality, feasible but suboptimal plans, subjective infeasibility, and incomplete structure are mutually exclusive. Each row contains 335 agent proposals. Recoverable fences and work-field packaging are excluded from the last category. Suboptimality means constraints are met but cost minimization fails; subjective infeasibility violates constraints known to the target agent.

\section{Matched-story Supplied-belief Controls}
\label{app:comparison}
\begin{figure}[!htb]
\centering
\includegraphics[width=\linewidth]{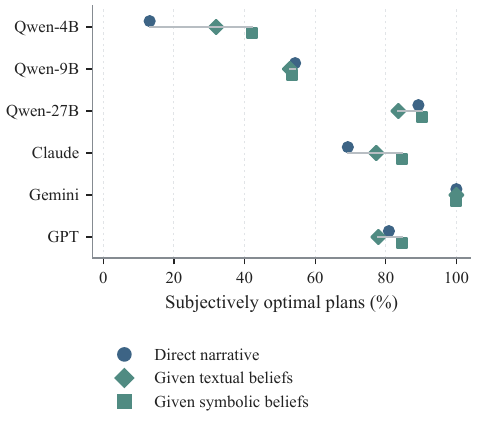}
\caption{Supplied-belief controls on the same 67 stories and 335 agent proposals. All conditions target complete subjectively optimal schedules.}
\label{fig:calbench-controls}
\end{figure}

Each main-table interface uses its common complete set across models. Table~\ref{tab:representation} and Figure~\ref{fig:calbench-controls} additionally hold 67 stories fixed across all three planning interfaces and six configurations. Textual and symbolic conditions both supply correct personal states but differ in representation and reading demands.

\begin{table*}[tp]
\centering\small
\begin{tabularx}{\textwidth}{@{}Yrrrrrr@{}}
\toprule
Metric & Qwen-4B & Qwen-9B & Qwen-27B & Claude & Gemini & GPT \\
\midrule
Direct narrative & 44/335 & 182/335 & 299/335 & 232/335 & 335/335 & 271/335 \\
Given textual beliefs & 107/335 & 177/335 & 280/335 & 259/335 & 335/335 & 261/335 \\
Given symbolic beliefs & 141/335 & 179/335 & 302/335 & 283/335 & 335/335 & 283/335 \\
\bottomrule
\end{tabularx}
\caption{Matched-story planning controls: 67 stories complete in all three interfaces and all six configurations, with five proposals per story.}
\label{tab:representation}
\end{table*}

\section{Joint Decisions, Routes, and Communication}
\begin{figure}[!htb]
\centering
\includegraphics[width=\linewidth]{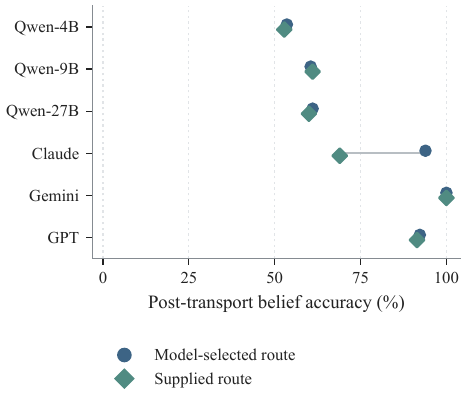}
\caption{Post-action belief accuracy, averaged equally over 35 common route configurations. Circles denote model-selected routes and diamonds supplied routes. Main-table entry accuracy instead pools all 474 entries.}
\label{fig:cwah-route}
\end{figure}

\begin{figure}[!htb]
\centering
\includegraphics[width=\linewidth]{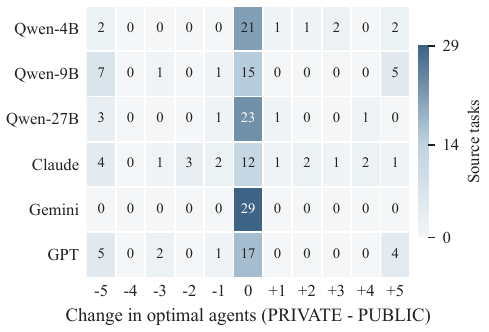}
\caption{PRIVATE minus PUBLIC in the number of optimal agents on 29 common source-task pairs. Cells count tasks. Zero change need not mean correctness. Agents are repeated measurements, not independent tasks.}
\label{fig:calbench-communication}
\end{figure}

Figure~\ref{fig:cwah-action} in the main text connects correct source choices to independent belief judgments on 218 nodes. Figures~\ref{fig:cwah-route} and~\ref{fig:calbench-communication} extend the analysis to post-action beliefs and communication conditions, using 35 matched route configurations and 29 complete source-task pairs. Each comparison retains its own outcome and statistical unit.
\section{Resources and Computational Scope}
\label{app:resources}
The experiments use the C-WAH distribution associated with \citet{zhang2024}, VirtualHome's Linux \texttt{v2.3.0} executable, and the public uniform/varied CalBench fixtures of \citet{zou2026}. The frozen releases used here are the household protocol dated October 5, 2026 and calendar story package dated October 9, 2026. Calendar selection uses 45 tasks from each public fixture, not all 180 upstream tasks. Prompts, source identifiers, seeds, reference states, semantic scorers, and native receipts identify the evaluated material.

The Qwen checkpoints are the unquantized BF16 releases \href{https://huggingface.co/Qwen/Qwen3.5-4B}{Qwen3.5-4B}, \href{https://huggingface.co/Qwen/Qwen3.5-9B}{Qwen3.5-9B}, and \href{https://huggingface.co/Qwen/Qwen3.5-27B}{Qwen3.5-27B}; all three model cards specify Apache-2.0. The 4B service uses one RTX 3090 and vLLM 0.26.0. The 9B and reused 27B services run on Ascend infrastructure. Native simulation runs on a Linux host with RTX A6000 hardware, using Python 3.8; the collection environment uses Python 3.11. No training is performed.

The logical evaluation budget is 3222 fixed requests and 158 native jobs per configuration, with up to 20 decisions per continuation. Actual calls depend on trajectories and bounded retries.

The task implementations are available from the \href{https://github.com/UMass-Embodied-AGI/CoELA}{C-WAH source repository} and the \href{https://github.com/bosonphoton/calbench2026}{CalBench source repository}.

\section{Task Construction and Node Sampling}
\label{app:construction}
\subsection{C-WAH Native Prefixes and Checkpoints}
The formal set retains 43 supported configurations across seven houses and their fixed seeds. Each configuration executes a native preparation trajectory of at most 120 steps. Failed-action feedback and actual states remain in the record. Nodes are selected at six event types: first possession, first placement, first incidental belief update, before a message, after a message, and a stable state. Only the first occurrence of each type is retained, with multiple types at one step merged. Sampling precedes model calls and yields 218 nodes, each with nine interfaces, totaling 1962 requests.

Continuation starts from the same preparation state within each configuration. Preparation and continuation share a 250-step budget, and the model coordinator makes at most 20 decisions. The two model conditions receive the full narrative with or without additional current beliefs. Belief-first, world-first, and exploration-first scripts provide executable strategy references. Action feedback, message receipts, and personal observations enter later history. Information obtained through exploration can change the next decision. Merely proposing exploration or a message does not yet change anyone's evidence.

\begin{table}[!ht]
\centering\small
\begin{tabularx}{\linewidth}{@{}Yrrr@{}}
\toprule
Script & Success/43 & Goals met & Steps \\
\midrule
Belief-first & 42 & 189/190 & 4710 \\
World-first & 13 & 122/190 & 4525 \\
Exploration-first & 39 & 186/190 & 5369 \\
\bottomrule
\end{tabularx}
\caption{Shared deterministic references. Each strategy uses the same tasks and native preparation states for 43 episodes. Step totals include preparation. Scripts are not counted again for each model configuration.}
\label{tab:script-reference}
\end{table}

The route experiment selects a held object with an unfinished goal at a native checkpoint. It compares a route proposed on the public house graph with a supplied graph route. Of the 43 configurations, six lack a public path connecting the held object's room to its goal room, and one lacks a valid held object with an unfinished goal. The 36-configuration route set is therefore fixed before model calls. Two conditions per configuration schedule 72 jobs and 108 logical model requests; accepted coverage is reported in Table~\ref{tab:native-all}.

Both conditions start from matched checkpoints: task, world relations, person rooms, relevant physical graph, belief values, and evidence timing must agree. Coordinate tolerance for people and the target object is 0.001. Native prefix replay must pass these and input-sufficiency checks before a model request. Each logical request adopts one complete answer. Six-model paired analysis uses the 35 configurations valid in both arms for every model. Supplied routes are shortest paths on the public graph. Incidental encounters and observations along them update later beliefs.

\subsection{CalBench Story Construction}
Frozen source files contain 45 uniform-cost and 45 tiered-cost tasks, with 30 each at easy, medium, and hard difficulty. Each retains five agents, sixteen slots, five meetings, native commitment costs, meeting participants, and its seed. Within each cost group, tasks are indexed by source-file order from $i=0$. Change owners are $i\bmod5$ and $(i+1)\bmod5$, and the intermediary is $(i+2)\bmod5$. Roles are fixed before model calls.

For each owner, immovable commitments in the final calendar are listed in slot order. We select index $\lfloor i/5\rfloor\bmod k$, where $k$ is the owner's number of immovable commitments. Initially, the story makes these two commitments movable, preserving their original costs and all other entries. Each owner later learns that the commitment must remain in its original slot and sends an update. The final world calendar equals the source task.

Events E01--E05 broadcast initial calendars. E06 and E08 are the two private changes, and E07 and E09 their messages. PUBLIC delivers both messages to all agents. PRIVATE delivers each only to its sender and the common intermediary. The variants share physical changes and differ in recipients. Each task has two stories, with independent requests for seven interfaces. In PRIVATE, the two owners and intermediary know different updates, while the other two agents retain their initial calendars.

\section{Native Evidence and Reference Belief Rules}
\label{app:gold}
\subsection{C-WAH Observations, Messages, and Search}
Personal observations are sampled only at recorded joint-step boundaries. A person occupies the room recorded at that boundary. Their native symbolic local view contains rooms and entities connected to it through INSIDE relations, except entities whose immediate INSIDE parent is not a room and lacks the OPEN state. Objects held by that person are included. The rule checks the immediate parent rather than recursively simulating visual occlusion.

A location relation supplies evidence only when both endpoints are visible. Possession takes precedence over storage, and storage uses the immediate parent rather than its room. Successful manipulation reveals the resulting object location to its actor. Co-presence in a room does not automatically reveal closed-container contents. ON and INSIDE are distinct. Physical feedback to the coordinator is not automatically delivered to executors, and imagined intermediate positions generate no observations.

Beliefs update through personal actions, actual observations, and delivered messages. A received assertion can induce a false belief. In the same exchange, direct observation overrides a conflicting message. A later assertion can update an older belief. A completed search that finds the object absent rules out the believed source. Without another learned source, belief becomes UNKNOWN. UNKNOWN means the source is undetermined, not that the object does not exist.

Access scoring retains each person's positive location assertions for each object, including superseded assertions, with duplicates removed. Negative search outcomes are excluded from this set. Belief is the current reference state after temporal updating. Access asks which location evidence the person has obtained, while belief asks where they now think the object is.

\subsection{CalBench Reference States and Subjective Planning}
The initial five broadcasts give everyone the five initial calendars. A personal change updates only its owner. Group messages update the corresponding entries for all recipients, while private messages update only sender and recipient. Messages are true when sent. Agents remember accurately and retain earlier values without new evidence. References follow chronological event replay.

CalBench access targets are ordered lists of original event identifiers. Sending, receiving, and personal changes count. A later relay does not additionally grant access to an original event the agent did not experience. These lists differ from C-WAH's deduplicated sets of location assertions, so access scores are not pooled across domains.

Supplied-belief conditions show actual and correct subjective calendars with their roles distinguished. Identical subjective calendars are grouped, allowing one solution to be reused for proposers with the same perspective. PUBLIC has one shared perspective, whereas PRIVATE may require several. The communication comparison therefore also changes the number of planning problems and reading demands.

\section{Formal Prompts and Output Protocol}
\label{app:prompts}
\subsection{CalBench Private-communication Prompt}
The following is an English translation of the complete SYSTEM and USER messages for DIRECT on tiered-cost task b032. The experiment used the Chinese original. Figure~\ref{fig:perspective} shows the actual scheduling response and native validation. Numerical values in the output template illustrate structure rather than the answer. Messages are grouped by role and JSON is indented for readability.

\begin{promptblock}{SYSTEM (English translation)}
Solve the problem step by step, strictly respecting its information boundaries. Return one JSON object with exactly two outer fields: work is a string array recording the calculations, and answer is the final answer object. Write work before answer. Do not add Markdown fences or text outside JSON. The field names, nesting, and value types in answer must match the template. Write arrays compactly.
\end{promptblock}

\begin{promptblock}{USER (English translation)}
There are five agents, A, B, C, D, and E. Each calendar has sixteen equal-length slots numbered 0--15. Each meeting occupies one slot.

A free slot costs 0. A movable commitment can be moved to another free slot in the same person's calendar at its specified one-time cost. An immovable commitment cannot be moved.

All five meetings below must be scheduled. Meetings sharing a participant cannot overlap. Meetings with disjoint participants may occupy the same slot.

Every agent must propose a complete schedule for all five meetings, minimizing the total additional displacement cost across participants under that agent's subjective calendars.

These are independent proposals, not an executed joint decision. Proposals may differ and do not update one another's beliefs.

The executor moves movable commitments occupying meeting slots to free slots that the person does not use for any meeting. Each calendar has enough initial free slots for these moves. The model only needs to choose meeting times.

Everyone immediately knows changes to their own calendar. Group messages reach all five agents. Private messages are visible only to their sender and specified recipient and are not automatically forwarded.

Messages are true and accurate when sent. Each person remembers prior information accurately and updates the same fact using later information. Without new information, prior judgments persist.

Your access to the complete material does not make every fact available to every agent. There are no changes or communications other than those specified.

Calendar output uses sixteen-entry arrays indexed by slots 0--15. Zero means free, a positive integer is the displacement cost, and the string "B" means immovable.

\medskip\textbf{All meetings}

Meeting 1: participants B, D, E.

Meeting 2: participants A, C, E.

Meeting 3: participants A, C, D.

Meeting 4: participants B, C, E.

Meeting 5: participants A, B, D.

\medskip\textbf{Complete story in chronological order}

E01. A says in the five-person group conversation: ``My calendar: slot 0 has a movable commitment costing 1; slot 1 has a movable commitment costing 100; slot 2 has a movable commitment costing 10; slot 3 is free; slot 4 has an immovable commitment; slot 5 is free; slot 6 has a movable commitment costing 1; slot 7 has an immovable commitment; slot 8 has a movable commitment costing 100; slot 9 has a movable commitment costing 1; slots 10 and 11 are free; slot 12 has a movable commitment costing 1; slot 13 has a movable commitment costing 10; slot 14 has a movable commitment costing 100; slot 15 is free.''

E02. B says in the five-person group conversation: ``My calendar: slot 0 has an immovable commitment; slot 1 has a movable commitment costing 10; slot 2 is free; slot 3 has an immovable commitment; slot 4 has a movable commitment costing 1; slot 5 is free; slot 6 has a movable commitment costing 10; slot 7 is free; slot 8 has a movable commitment costing 1; slot 9 is free; slot 10 has an immovable commitment; slot 11 has a movable commitment costing 10; slots 12, 13, and 14 are free; slot 15 has a movable commitment costing 1.''

E03. C says in the five-person group conversation: ``My calendar: slot 0 has a movable commitment costing 1; slot 1 has an immovable commitment; slot 2 is free; slot 3 has a movable commitment costing 1; slot 4 has a movable commitment costing 100; slots 5 and 6 have immovable commitments; slot 7 is free; slot 8 has a movable commitment costing 10; slot 9 has a movable commitment costing 100; slot 10 has a movable commitment costing 10; slot 11 has a movable commitment costing 100; slot 12 has a movable commitment costing 1; slot 13 is free; slots 14 and 15 have immovable commitments.''

E04. D says in the five-person group conversation: ``My calendar: slot 0 has an immovable commitment; slot 1 has a movable commitment costing 10; slot 2 is free; slot 3 has an immovable commitment; slots 4 and 5 are free; slot 6 has a movable commitment costing 100; slots 7 and 8 are free; slot 9 has a movable commitment costing 100; slot 10 has a movable commitment costing 1; slot 11 has an immovable commitment; slots 12 and 13 are free; slot 14 has an immovable commitment; slot 15 has a movable commitment costing 100.''

E05. E says in the five-person group conversation: ``My calendar: slot 0 has an immovable commitment; slots 1 and 2 have movable commitments costing 10; slot 3 is free; slots 4 and 5 have movable commitments costing 1; slot 6 is free; slot 7 has a movable commitment costing 10; slots 8 and 9 are free; slot 10 has an immovable commitment; slots 11 and 12 have movable commitments costing 1; slot 13 has a movable commitment costing 10; slot 14 is free; slot 15 has a movable commitment costing 100.''

E06. B receives notice that the slot-8 commitment must occur at its original time and is now immovable. Other slots are unchanged.

E07. B privately tells D: ``My slot-8 commitment is now immovable. Other slots are unchanged.''

E08. C receives notice that the slot-0 commitment must occur at its original time and is now immovable. Other slots are unchanged.

E09. C privately tells D: ``My slot-0 commitment is now immovable. Other slots are unchanged.''

\medskip\textbf{Question}

For each agent, solve for the minimum-total-cost schedule of all five meetings under that agent's subjective calendars. Any tied optimum is acceptable. The final answer object maps A--E to objects mapping meeting identifiers "1"--"5" to integer slots 0--15.

\medskip\textbf{Joint scheduling method}

Every meeting has three participants, and any pair of meetings shares at least one. Thus, in any one agent's proposal, the five slots for meetings 1--5 must be distinct. Different agents' independent proposals can use the same slots.

Solve separately for each proposer using only the five calendars that proposer believes. Do not overwrite unknown changes with the true calendar.

1. For every meeting $m$ and slot $s$, inspect each of the three participants. If any entry is "B", the candidate is infeasible. Otherwise, candidate cost $c(m,s)$ is the sum of the three entries, with free slots costing 0.

2. Jointly select five distinct slots to minimize the sum of $c(m,s)$ across meetings. Do not choose each meeting's cheapest slot independently and simply combine them. For conflicts, compare the total costs of moving different meetings to alternatives, including changes to several meetings at once.

3. First obtain a feasible assignment with five distinct slots, then compare alternatives. The sum of independent minimum meeting costs is a lower bound. A conflict-free assignment attaining it is optimal. Otherwise compare the extra joint cost induced by conflicts. Do not treat the first feasible assignment as optimal.

4. Before output, verify that each proposal contains exactly meetings 1--5, that the five values are distinct, and that no participant has an immovable commitment at their meeting time. Check the total cost. Proposers with identical subjective calendars may reuse one complete optimal schedule.

\medskip\textbf{Independent teaching example, not this task's calendars or answer}

Three mutually conflicting meetings X, Y, Z have the following total displacement costs at candidate slots 0, 1, 2, 3:

X: [0, 1, 9, 9]; Y: [0, 5, 9, 9]; Z: [9, 9, 0, 2].

Putting both X and Y at 0 is invalid. X=0, Y=1, Z=2 is valid but costs 5.

X=1, Y=0, Z=2 costs 1 and is optimal. X and Y cannot both use 0, so at least one must pay at least 1, and this assignment attains that bound.

Recompute using this task's calendars. Do not copy the teaching example's slots.

In work, write the actual calculations in this order: establish the differences among subjective calendars; for each distinct subjective calendar, list feasible candidates for meetings 1--5 as (slot: sum of the three participant costs), excluding infeasible slots; compare joint assignments to five distinct times and give the chosen total cost. Do not merely claim that a plan was checked or is optimal. Compute each identical subjective calendar once and reuse it across proposers. Do not enumerate every permutation. Prioritize conflicts and reassignment among low-cost candidates.

\medskip\textbf{Complete output structure template}

{\ttfamily\raggedright
\noindent\{\par
\noindent\hspace*{.9em}"work": [\par
\noindent\hspace*{1.8em}"Record derivations and calculations here"\par
\noindent\hspace*{.9em}],\par
\noindent\hspace*{.9em}"answer": \{\par
\noindent\hspace*{1.8em}"A": \{"1": 0, "2": 1, "3": 2,\par
\noindent\hspace*{2.7em}"4": 3, "5": 4\},\par
\noindent\hspace*{1.8em}"B": \{"1": 0, "2": 1, "3": 2,\par
\noindent\hspace*{2.7em}"4": 3, "5": 4\},\par
\noindent\hspace*{1.8em}"C": \{"1": 0, "2": 1, "3": 2,\par
\noindent\hspace*{2.7em}"4": 3, "5": 4\},\par
\noindent\hspace*{1.8em}"D": \{"1": 0, "2": 1, "3": 2,\par
\noindent\hspace*{2.7em}"4": 3, "5": 4\},\par
\noindent\hspace*{1.8em}"E": \{"1": 0, "2": 1, "3": 2,\par
\noindent\hspace*{2.7em}"4": 3, "5": 4\}\par
\noindent\hspace*{.9em}\}\par
\noindent\}\par
}
The template illustrates structure and types only. Its numerical examples are not this task's answer. Solve again and fill in the actual values. The only outer fields are work and answer, containing derivations and the answer, respectively. Calendar arrays must have sixteen entries. Meeting slots are integers, such as 3 rather than "3". Do not add costs, explanations, or other fields to answer.
\end{promptblock}

\subsection{C-WAH Public Rules and Teaching Example}
Houses, objects, and complete physical histories vary with the frozen node. The following reproduces the shared English SYSTEM rules and complete teaching dialogue of formal WORLD\_ACCESS requests. Node histories contain successive physical boundaries, supplied as described in the main text.

\begin{promptblock}{SYSTEM}
You read a household story about Alice and Bob. The narrator can know physical facts that neither person has learned. Each person uses only their own successful actions,\allowbreak{} actual observations,\allowbreak{} and messages they actually receive. Nobody has additional object-location memories before the opening story. Moving an object does not update an absent person's belief. A received assertion can create a false belief. Direct observation takes precedence over a conflicting message during the same exchange; a later delivered assertion can update an older belief. A complete check that rules out the believed source leaves UNKNOWN unless another source has been learned. UNKNOWN is not absence of the object.

ASSIGN describes a plan using the assigned person's existing belief,\allowbreak{} not a transmission of new location information. The source must equal that person's current believed location. SEND communicates a location assertion supported by the sender's belief. A proposed or future communication has not updated the recipient yet. Cognition-consistent plans are not necessarily efficient or physically successful.

Return only the requested JSON,\allowbreak{} with the supplied semantic identifiers. Do not explain your answer.

The initial map and each physical report are coordinator information,\allowbreak{} not messages to the people. Personal evidence is sampled only at the recorded joint step boundaries,\allowbreak{} not at imagined intermediate positions. At a boundary each person occupies their recorded room. In the native symbolic observation rule,\allowbreak{} their local view includes all rooms and entities connected through INSIDE relations to their current room,\allowbreak{} except an entity whose immediate INSIDE parent is not a room and does not have state OPEN. Objects held by that person are also included. The rule checks the immediate parent,\allowbreak{} not photograph-level occlusion or recursive visual geometry. A location relation supplies evidence only when both its endpoints are included in that local view. Held-by relations take precedence over storage relations; for storage,\allowbreak{} the immediate parent takes precedence over its room. A successful manipulation supplies its actor with the resulting item location. Room co-presence alone does not reveal the contents of a closed container. An ON relation and an INSIDE relation are different. Only actual delivered message receipts update recipients. Unmentioned physical relations persist until removed or changed. A completed search rules out that searched source when its physical item is elsewhere. No one has extra pre-story memories.
\end{promptblock}

\begin{promptblock}{USER}
The library and study are connected. Alice and Bob initially occupy the library. book\_\allowbreak{}1 is ON shelf and INSIDE library; shelf is INSIDE library; bench is INSIDE study. At boundary 1 Alice is in study and Bob remains in library. At boundary 2 Bob successfully grabs book\_\allowbreak{}1 in library. At boundary 3 Bob and book\_\allowbreak{}1 are in study with Alice; Bob holds book\_\allowbreak{}1. At boundary 4 Alice returns to library; Bob remains in study holding book\_\allowbreak{}1. At boundary 5 Bob successfully puts book\_\allowbreak{}1 ON bench and INSIDE study; Alice stays in library.

Return \{"world":\allowbreak{}\{item:\allowbreak{}location\},\allowbreak{}"access":\allowbreak{}\{"Alice":\allowbreak{}\{item:\allowbreak{}[locations]\},\allowbreak{}"Bob":\allowbreak{}\{item:\allowbreak{}[locations]\}\}\}. Include every listed item. Access lists all positive location assertions personally obtained by observation,\allowbreak{} own action,\allowbreak{} or received message,\allowbreak{} including superseded assertions. Do not include narrator-only facts or negative checks; deduplicate each list.
\end{promptblock}

\begin{promptblock}{ASSISTANT}
{\ttfamily\raggedright
\noindent\{\par
\noindent\hspace*{.9em}"world": \{"book\_1": "bench"\},\par
\noindent\hspace*{.9em}"access": \{\par
\noindent\hspace*{1.8em}"Alice": \{\par
\noindent\hspace*{2.7em}"book\_1": [\par
\noindent\hspace*{3.6em}"shelf", "carried\_by\_Bob"\par
\noindent\hspace*{2.7em}]\par
\noindent\hspace*{1.8em}\},\par
\noindent\hspace*{1.8em}"Bob": \{\par
\noindent\hspace*{2.7em}"book\_1": [\par
\noindent\hspace*{3.6em}"shelf",\par
\noindent\hspace*{3.6em}"carried\_by\_Bob", "bench"\par
\noindent\hspace*{2.7em}]\par
\noindent\hspace*{1.8em}\}\par
\noindent\hspace*{.9em}\}\par
\noindent\}\par
}
\end{promptblock}

\subsection{Interface Output Objects}
All CalBench interfaces use an outer \texttt{work} string array and \texttt{answer} object. World/access answers contain \texttt{world} and \texttt{access}. The two belief interfaces contain only \texttt{beliefs}. Supplied access contains both \texttt{beliefs} and \texttt{proposals}. Other planning answers directly map each of the five agents to a proposal.

Calendars use 0, positive integer costs, or the string B, in arrays of exactly sixteen entries. Beliefs are nested by holder and calendar owner, with five arrays per holder. Proposals are nested by proposer and meeting identifiers ``1''--``5'', with integer slots 0--15. The work field expresses derivations. Content scoring uses the final answer.

C-WAH beliefs specify Alice's and Bob's believed sources for all target objects. Fixed ASSIGN answers preserve the specified actor, object, and goal, supplying only the believed source. Free continuation returns a \texttt{commands} array permitting ASSIGN, SEND, and EXPLORE. Commands for one person execute sequentially, and the two queues run concurrently. Cognitive checks use the state before submission of the entire command group. Undelivered messages cannot justify assignments in that group that depend on them.

\section{Scoring and Aggregation}
\label{app:scoring}
\subsection{Strict State and Plan Metrics}
Let $\mathcal C$ denote scored factual entries for an interface. Entry accuracy is
\begin{equation}
A_{\rm cell}=\frac{1}{|\mathcal C|}\sum_{c\in\mathcal C}\ind[\hat y_c=y_c].
\end{equation}
A CalBench agent belief is correct only if all 80 entries in its five calendars match. Complete-state accuracy is
\begin{equation}
A_{\rm agent}=\frac{1}{|\mathcal A|}\sum_{a\in\mathcal A}\prod_{c\in\mathcal C_a}\ind[\hat y_c=y_c],
\end{equation}
where $\mathcal A$ is the set of scored agent instances in complete responses and $\mathcal C_a$ its 80 entries. C-WAH access uses exact set matching per person and object, while beliefs use entry matching. A missing extractable target receives zero for the relevant interface metric. Recoverable code fences and surrounding text do not affect scoring.

Each complete CalBench proposal is passed to the native executor. It checks meeting identifiers, integer slots, participant conflicts, and immovable commitments. Movable commitments occupying meeting slots are relocated to free positions outside all of that person's reserved meeting slots, with every action validated. Feasible plans are optimal when their cost equals the frozen subjective minimum. All tied optima are accepted. The same proposal is separately validated in the actual calendar for world feasibility.

Because every pair of meetings shares a participant, optimization reduces to assigning different meetings to different slots. Let $P_m$ be the participants in meeting $m$, and $c_a(m,s)$ the sum of their displacement costs at slot $s$ as believed by proposer $a$. A candidate involving an immovable commitment is disabled. We solve
\begin{align}
\min_z\quad &\sum_{m=1}^{5}\sum_{s=0}^{15}c_a(m,s)z_{ms},\\
\text{s.t.}\quad &\sum_s z_{ms}=1,\quad\sum_m z_{ms}\leq1,\\
&z_{ms}\in\{0,1\},\\
&z_{ms}=0\quad\text{for disabled candidates}.
\end{align}
The reference optimum is established when CP-SAT, independent assignment optimization, and native replay agree. Task construction verifies sufficient free destination slots.

\subsection{Response Selection and Task-level Aggregation}
A complete output requires \texttt{finish\_reason=stop}. An incomplete request permits at most three additional attempts, stopping at the first complete output. A prespecified transport-only supplement allows three further attempts for 61 affected 4B fixed keys, 110 9B fixed keys, and affected decisions in 20 9B native jobs, with absolute attempt counts preserved. A new truncation ends that supplement. Correctness and scorability never determine selection. The common extractor preserves genuinely missing fields, wrong states, infeasibility, and suboptimality. Six-model coverage is reported in Tables~\ref{tab:full-coverage} and~\ref{tab:interface-coverage}.

Scoring proceeds by (1) identifying the unique adopted response by story and interface, (2) extracting its existing target, (3) comparing world, access, and belief references, (4) replaying each proposal and comparing subjective cost and world feasibility, and (5) aggregating PUBLIC and PRIVATE by source task. Both conditions must be complete for paired analysis. Complete wrong answers remain in denominators. Interface comparisons match stories without selecting on correctness.

C-WAH nodes are repeated observations within configurations. CalBench agents and communication stories repeat within source tasks. Configuration averages first compute accuracy within each configuration and then weight configurations equally. Pooled-entry main-table accuracy divides all correct entries by all scored entries. These weights are not interchanged. We report raw counts, paired differences, and single-run configuration results, without treating repeated observations as independent trials for significance claims.

\section{An Actual Case: Beliefs and Task Verification}
\label{app:case}
This section presents actual Qwen3.5-27B outputs for the PRIVATE story of tiered-cost source task b032: a structured-belief response and the complete proposals in Figure~\ref{fig:perspective}. The initial broadcast says B's slot-8 commitment is movable at cost 1. E06 states that B learns it must remain in its original slot. E07 is B's private message only to D: ``My slot-8 commitment is now immovable. Other slots are unchanged.'' A, C, and E do not receive it. The other change makes C's slot-0 commitment immovable, also reported only to D.

\subsection{Information Attribution in Belief Answers}
\begin{table}[ht]\centering\small
\begin{tabularx}{\linewidth}{@{}Yrrl@{}}\toprule
Holder & Reference & Model & Update received \\
\midrule
A & 1 & B & None \\
B & B & B & Own change \\
C & 1 & B & None \\
D & B & B & Private message \\
E & 1 & B & None \\
\bottomrule\end{tabularx}
\caption{Five agents' beliefs about B's slot 8 in the actual case. The number 1 is a displacement cost, and the string B means immovable. Model values come directly from the complete structured response.}
\label{tab:real-case}\end{table}

\paragraph{Actual output excerpt.}
The following preserves the model's complete sixteen-entry calendar for A's belief about B. Positions are zero-indexed.

\begin{promptblock}{BELIEF\_STRUCT: output excerpt}
{\ttfamily\raggedright
\noindent\{"A": \{"B": [\par
\noindent\hspace*{.9em}"B", 10, 0, "B", 1, 0, 10, 0,\par
\noindent\hspace*{.9em}"B", 0, "B", 10, 0, 0, 0, 1\par
\noindent]\}\}\par
}
\end{promptblock}

A obtains neither the E06 personal change nor the E07 private message, so slot 8 should remain 1 in A's reference calendar. The output replaces this entry with world truth for all three uninformed agents. It gets 397 of 400 entries correct, but only B and D have fully correct eighty-entry states. Entry accuracy and complete-perspective accuracy thus provide different diagnoses.

\subsection{Complete Plans and Native Verification}
DIRECT independently reads the same story and schedules all five meetings for each agent. Participants in meetings 1--5 are BDE, ACE, ACD, BCE, and ABD. Table~\ref{tab:case-plans} gives the adopted complete output and validation from recorded native receipts.

\begin{table}[ht]\centering\small
\setlength{\tabcolsep}{3pt}
\begin{tabular}{@{}clrrcc@{}}\toprule
Agent & Meeting slots & Cost & Opt. & \shortstack{Belief\\feasible} & \shortstack{World\\feasible} \\
\midrule
A & 8,3,12,2,5 & 14 & 14 & Yes & No \\
B & 8,3,12,2,5 & --- & 15 & No & No \\
C & 8,3,12,2,5 & 14 & 14 & Yes & No \\
D & 12,3,2,7,5 & 22 & 15 & Yes & Yes \\
E & 8,3,12,2,5 & 14 & 14 & Yes & No \\
\bottomrule\end{tabular}
\caption{Complete proposals for b032 PRIVATE. Slots correspond to meetings 1--5. Cost is total displacement under the subjective calendar, and Opt. is the reference minimum.}
\label{tab:case-plans}\end{table}

A, C, and E propose cost-14 plans that attain their subjective optima. B's identical proposal violates the known slot-8 constraint. D's plan is feasible but costs 22 rather than the minimum 15. The case separates subjective infeasibility from feasible but suboptimal planning.

Figure~\ref{fig:perspective} highlights A and B. Under A's calendars, native replay first moves B's commitment from slot 8 to 7, then schedules meeting 1 at 8. All five meetings require five commitment moves at total cost 14. Under B's reference calendar, the slot-8 commitment is immovable and this action is rejected. Both proposals fail in the actual world, but A's remains optimal under A's information.

Structured-belief and direct-planning responses are independent requests at the same node. The former answer is not passed to the latter. This example distinguishes belief answers, perspective-specific planning, and actual execution conditions.

\section{Scoring Granularity and a Deterministic Reference}
\label{app:representation}
\subsection{Copying the World Does Not Establish Perspective Recovery}
Assign every agent the actual final calendar as its predicted belief and score against the frozen references. The 180 stories have 72000 belief entries: 71460 agree with the world, while 540 should retain earlier values. Copying scores 99.25\% by entry but only 540/900 complete agent states: 450/450 in PUBLIC and 90/450 in PRIVATE.

In PRIVATE, only the intermediary receives both updates. Each change owner lacks the other's update, and the remaining two agents lack both. Copying the world therefore gets only the intermediary's complete state correct in each task. Shared facts dominate entry accuracy, whereas complete personal states and stale entries directly test information boundaries. This deterministic reference follows reference states, not an additional model or model output.

\begin{figure}[!htb]
\centering
\includegraphics[width=\linewidth]{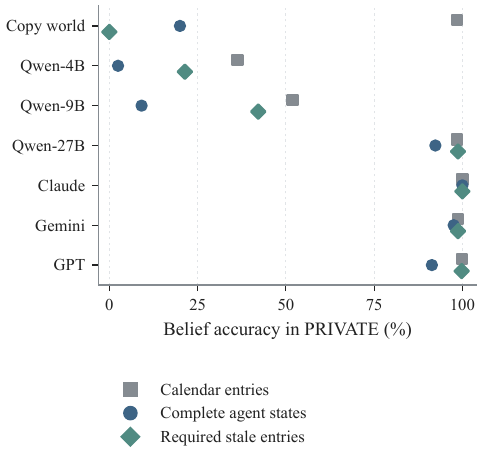}
\caption{Belief granularity on 81 common PRIVATE stories: 32400 entries, 405 complete agent states, and 486 stale entries. Copying the world scores 98.5\%, 20.0\%, and zero, respectively. The construction-wide counts in the text use all 90 source tasks.}
\label{fig:calbench-metrics}
\end{figure}

\subsection{World Plans and Perspective-sensitive Planning}
\label{app:planning-overlap}
For each frozen story, we take its solver-produced world-optimal plan and evaluate that same plan against every agent's subjective calendars using the original feasibility and optimality checks. Across all 180 stories, it is subjectively feasible in 900/900 cases and optimal in 863/900 (95.9\%). On PRIVATE stories, it is optimal in 413/450 (91.8\%). These counts concern plan correctness, not recovery of the full personal state.

On the six-model common DIRECT support (97 stories), the world plan is optimal in 466/485 cases. In the remaining 19 cases from six source tasks, subjective optimal cost is strictly below world-optimal cost, so the world plan cannot satisfy subjective optimality. Table~\ref{tab:strict-planning} reports models on this strict-cost-separation subset. The subset diagnoses where this particular world-plan reference fails; it does not establish that every successful plan requires an explicit complete belief representation. Source tasks remain the clustering unit.

\begin{table}[htb]
\centering\small
\begin{tabular}{@{}lr@{}}
\toprule
Configuration & Optimal proposals \\
\midrule
Qwen3.5-4B & 0/19 \\
Qwen3.5-9B & 0/19 \\
Qwen3.5-27B & 13/19 \\
Claude Opus 5.5 & 10/19 \\
Gemini 3.1 Pro Preview & 19/19 \\
GPT-6.1 Sol & 10/19 \\
\bottomrule
\end{tabular}
\caption{Subjective optimality on 19 strict-cost-separation agent cases from six source tasks within common DIRECT support. All cases retain their original task constraints and scoring.}
\label{tab:strict-planning}
\end{table}

\subsection{State Representation and Planning}
The main table reports narrative and structured beliefs on each interface's common complete requests. Their complete-agent metric requires every calendar entry, without excluding recoverable JSON packaging. Because calendar coverage differs by interface, these columns are descriptive, not matched estimates of a representation effect. Household narrative and structured conditions have identical full coverage and permit a same-target comparison.

Table~\ref{tab:representation} compares textual and symbolic supplied beliefs on matched stories. Both provide correct states, leaving the need to construct each agent's subjective feasible set and minimize cost. Differences concern representation and downstream use, not direct measurements of internal reasoning mechanisms.

\end{document}